%% file: LumiSculpt.tex
\documentclass[sigconf, screen, nonacm]{acmart}

\AtBeginDocument{%
  }

\input{macros}

\usepackage{multirow}
\usepackage{xcolor}
\usepackage{graphicx}
\usepackage{amsmath}

\usepackage{amssymb}
\usepackage{booktabs}
\usepackage{subcaption}
\usepackage{wrapfig}
\usepackage{bm}
\usepackage{bbm}
\usepackage{tikz}
\usepackage{booktabs} 
\usepackage{lipsum}
\usepackage{colortbl}
\usepackage{bbding}
\usepackage{tabularx} 
\newcommand{\blfootnote}[1]{\begingroup
\renewcommand\thefootnote{}\footnote{#1}\addtocounter{footnote}{-1}
\endgroup}

\begin{document}

\title{LumiSculpt: Enabling Consistent Portrait Lighting in Video Generation}

\author{%
  Yuxin Zhang$^{1*}$,\space\space
  Dandan Zheng$^{2}$,\space\space
  Biao Gong$^{2\dagger}$,\space\space  
  Shiwen Wang$^{1}$,\space\space
  \\
  \vspace{1.2mm}
  Jingdong Chen$^{2}$,\space\space
  Ming Yang$^{2}$,\space\space
  Weiming Dong$^{1}$\raisebox{2pt}{\tiny{\Envelope}},\space\space
  Changsheng Xu$^{1}$
  \\
  \vspace{1mm}
  $^{1}$\normalfont{MAIS, Institute of Automation, Chinese Academy of Sciecnes}\quad
  $^{2}$\normalfont{Ant Group}\quad
  \\
  \tt\small 
  \{zhangyuxin2020, weiming.dong, changsheng.xu\}@ia.ac.cn
  \\
  \tt\small
  \{yuandan.zdd,gongbiao.gb,jingdongchen.cjd,m.yang\}@antgroup.com
}
\renewcommand{\shortauthors}{Zhang et al.}

\begin{abstract}
Lighting plays a pivotal role in ensuring the naturalness and aesthetic quality of video generation. However, the impact of lighting is deeply coupled with other factors of videos, \emph{e.g.}, objects and scenes. Thus, it remains challenging to disentangle and model coherent lighting conditions independently, limiting the flexibility to control lighting in video generation.  
In this paper, inspired by the established controllable T2I models, we propose \sysname, which enables precise and consistent lighting control in T2V generation models.
\sysname equips the video generation with new interactive capabilities, allowing the input of reference image sequences with customized lighting conditions. 
Furthermore, the core learnable plug-and-play module of \sysname facilitates direct control over the intensity, position and trajectory of an assumed light source in video diffusion models.
To effectively train \sysname and address the issue of insufficient lighting data, we construct \dataname, a new lightweight and flexible dataset for portrait lighting of images and videos. 
Experimental results demonstrate that \sysname achieves precise and high-quality lighting control in video generation.
The analysis demonstrates the flexibility of \dataname.
\end{abstract}



\begin{teaserfigure}
    \centering
    \includegraphics[width=0.99\linewidth]{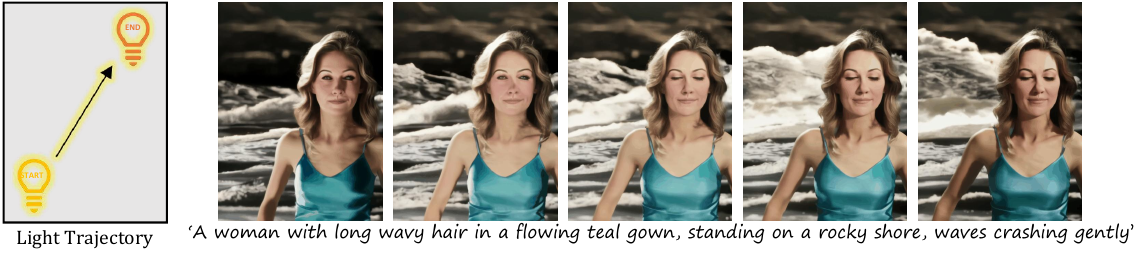}
    \caption{\sysname allows user-specified control over the intensity, position, and trajectories of an assumed light source, with textual conditions as input. Being trained once, \sysname is capable of generating diverse results at inference time.
}
\label{fig:teaser}
\end{teaserfigure}

\maketitle
\blfootnote{$*$ Work done during internship at Ant Group.\space\space $\dagger$\space Project lead.}
\blfootnote{{\tiny{\Envelope}} \space Corresponding author.}

\input{Sections/1-Intro}
\input{Sections/2-Related}
\input{Sections/3-Method}

\input{Sections/4-Exp}

\input{Sections/5-Conclusion}

\bibliographystyle{ACM-Reference-Format}
\bibliography{Light_main}

\end{document}

%% file: macros.tex
\newcommand{\sysname}{\textit{LumiSculpt}~}
\newcommand{\dataname}{\textit{LumiHuman}~}
\newcommand{\loss}{\mathcal{L}}

\definecolor{rebuttal}{rgb}{0,0,0}
\newcommand{\rebuttal}[1]{{\color{rebuttal}#1}}

%% file: Sections/1-Intro.tex
\section{Introduction}

If a video tells a story, then lighting is the voice that shapes its tone and mood.
Lighting is essential for video generation, which is one of the defining factors for the overall aesthetic quality of the generated video, and is also used to convey emotions, highlight character traits, and guide the audience's attention.
Mainstream video generation methods currently employ latent diffusion models (LDMs) to achieve video generation through multi-step denoising in a latent space.
Research on controllable image and video generation based on LDMs supports our studies of consistent lighting control.
Several methods~\cite{vdm,chen2023videocrafter1,imagenvideo,ModelScopeT2V,animatediff} have been developed to achieve relatively accurate text-controlled video generation, as well as video editing~\cite{chai2023stablevideo,ceylan2023pix2video}, customization~\cite{tuneavideo}, and controlling~\cite{wang2023videocomposer,wei2024dreamvideo,he2024cameractrl}. These works have improved the controllability, aesthetics, and usability of video generation. 
However, due to the deep coupling between lighting and the other non-stationary factors of videos, it is challenging to model coherent lighting conditions independently, resulting in a lack of handy approaches to controlling lighting in videos.

The challenge of customizing lighting lies in three aspects: the lack of training data, the representation of lighting, and the mechanism of injecting lighting conditions without influencing other attributes. 
Specifically, although there are relighting datasets based on light stages~\cite{lightfield}, the data format of light stages is not readily applicable in video generation scenarios. Therefore, a new dataset that is adaptable to text-controlled content generation is needed. 
Obtaining the projection of lighting on the camera's imaging plane requires knowledge of the lighting and the surface texture of the illuminated objects~\cite{kim2024switchlight,ren2024relightful,mei2024holo}, which cannot be achieved in an end-to-end video generation scenario. Thus, a simple lighting representation that is only related to lighting parameters is important. 
Finally, similar to most control tasks, lighting control faces the problem of the deep decoupling of lighting from other factors, such as semantics and color.

In this paper, we propose \dataname to address the problem of limited training data. \dataname is a portrait lighting dataset that can constitut more than 220K videos of humans with known lighting parameters. This is a lightweight and flexible dataset that is not limited to specific lighting movements but is presented in freely combinable frames, laying the foundation for a more diverse range of lighting paths and combinations. We then use virtual engine rendering with known lighting parameters to obtain projections of different directional lighting on planes as a lighting representation.
\rebuttal{To achieve video lighting control, we propose a lighting control method, \sysname, which learns an accurate plug-and-play lighting module capable of controlling the direction and movement of lighting in video generation. }
To solve the problem of lighting feature injection, we introduce a light control module that takes the lighting projection as input and integrates lighting control injected into the generative model layer by layer.
Furthermore, to better decouple lighting from appearance, we design a decoupling loss based on a dual-branch structure, preserving diverse generative capabilities.

We implemented \sysname\ on Open-Sora-Plan~\cite{opensora} to enable precise lighting control. 
\rebuttal{\sysname focuses specifically on lighting control in text-to-video generation, rather than addressing image-to-image or video-to-video relighting tasks. It goes beyond managing fixed types and directions of lighting based on text prompts. Instead, it enables precise control over the position and moving trajectory of the light source within a single video, offering great flexibility and customization.
}
We conducted comprehensive quantitative and qualitative evaluations. The experimental results show that \sysname\ has achieved state-of-the-art performance in the control of text-to-video lighting in precision and diversity, as shown in Figure~\ref{fig:teaser}. In summary, our main contributions are as follows:

\begin{itemize}
\item We introduce a portrait lighting dataset \dataname. \dataname is a lighting video dataset comprising over 220K different videos (\textit{i.e.,} 2.3M images). \dataname includes over 30K lighting positions, and over 3K light source trajectories for each individual.
\dataname paves the way for lighting control in both image and video generation.

\item We introduce \sysname, which enables control of the position and moving trajectories of light sources in video generation. We propose a lighting representation method, a lighting injection approach, and a lighting decoupling loss for the text-to-video generation scenario, which enables diverse content generation with limited data.

\item Extensive experiments prove that \sysname has achieved state-of-the-art performance. \sysname can enable consistent portrait lighting in diverse scenarios.
The analysis demonstrate the comprehensiveness and flexibility of \dataname, which provides a generalizable lighting prior.

\end{itemize}

%% file: Sections/2-Related.tex
\section{Related Works}

\subsection{Relighting}
In recent years, deep learning techniques have made significant progress in portrait relighting~\cite{kim2024switchlight, mei2023lightpainter, mei2024holo, nestmeyer2020learning, pandey2021total, sun2019single, wang2020single, yeh2022learning,  zhang2021neural}, which often rely on paired data captured by light stage systems~\cite{lightfield} for supervised learning. Typically, these methods require the use of high dynamic range (HDR) environmental maps as input. This process involves estimating intermediate surface properties, including normal vectors, albedo, diffuse reflectance, and specular reflection characteristics.  However, the reliance on HDR environmental maps limits the practical application of these techniques in video generation scenarios.
Besides, researchers also explore portrait relighting techniques that do not depend on light stage data~\cite{hou2021towards, hou2022face, wang2023sunstage}.

Recently, diffusion-based models have shed light on new approaches to relighting. Ren et al.~\cite{ren2024relightful} propose a three-stage lighting-aware diffusion model called Relightful Harmonization, which aims to provide complex lighting coordination for foreground portraits with any background image. Zeng et al.~\cite{zeng2024dilightnet} propose a three-stage portrait relighting method using a fine diffusion model called DiLightNet, which calculates radiance cues to re-synthesize and refine the foreground object by combining the rough shape of the foreground object inferred from the preliminary image. Xing et al.~\cite{xing2024retinex} propose a natural image relighting method called Retinex-Diffusion, which treats the diffusion model as a black-box image renderer and decomposes its energy function to be consistent with the image formation model. However, there still lack of methods for lighting control in text-to-video generation.
\rebuttal{The most related works are diffusion-based relighting methods. 
LightIt~\cite{lighit} is an image-guided method for image relighting conditioning on shading estimation and normal maps.
IC-Light~\cite{iclight} is an image relighting method to generate harmonized background with the user input foreground. 
Light-A-Video~\cite{zhou2025light} proposes a training-free video relighting method, which leverages on the image relighting model to achieve video relighting controlled by text.
It is worth noting that the above methods are image-to-image or video-to-video methods that can leverage the input prior, while our method aims at lighting control in text-to-video generation, which is more challenging. Moreover, our method is able to control the exact position and trajectory of the light source within a single video, instead of setting a fixed light type using text prompts.
}

\input{Figs/fig_indivisuals}

\input{Tables/tab_lumihuman}

\subsection{Text-to-video synthesis and controlling}
Recently, several researches, such as \cite{vdm,chen2023videocrafter1,imagenvideo,ModelScopeT2V,animatediff}, have adopted diffusion models to create highly realistic video content, utilizing text as conditions in guiding the generation process. These studies focus on ensuring alignment between textual descriptions and the final video output. 
Addressing the issue of difficulty in precisely describing specific visual attributes through text conditions, some studies have attempted to achieve finer video control by fine-tuning models or introducing additional control parameters. 
Tune-A-Video~\cite{tuneavideo} propose a fine-tuning framework that allows users to customize specific videos. 
VideoComposer~\cite{wang2023videocomposer} use explicit control signals to guide the temporal dynamics of the video.
Gong et al.~\cite{gong2023talecrafter} introduce TaleCrafter to handle interactions among multiple characters, featuring layout and structural editing capabilities. 
He et al.~\cite{he2023animate} propose a retrieval-based deep guidance method that can integrate existing video clips into a coherent narrative video by customizing the appearance of characters.
These studies mainly focus on the appearance of objects and scenes. 
Several methods~\cite{motiondirector,zhang2023motioncrafter,wei2024dreamvideo,he2024cameractrl} learn and control motion through customized diffusion models. These attempts have made substantial progress in controlling video in specific aspects. 
Nevertheless, further dedicated efforts are required on precise control of lighting in videos.

%% file: Figs/fig_indivisuals.tex
\begin{figure*}[t]
\begin{center}
\includegraphics[width=0.9\linewidth]{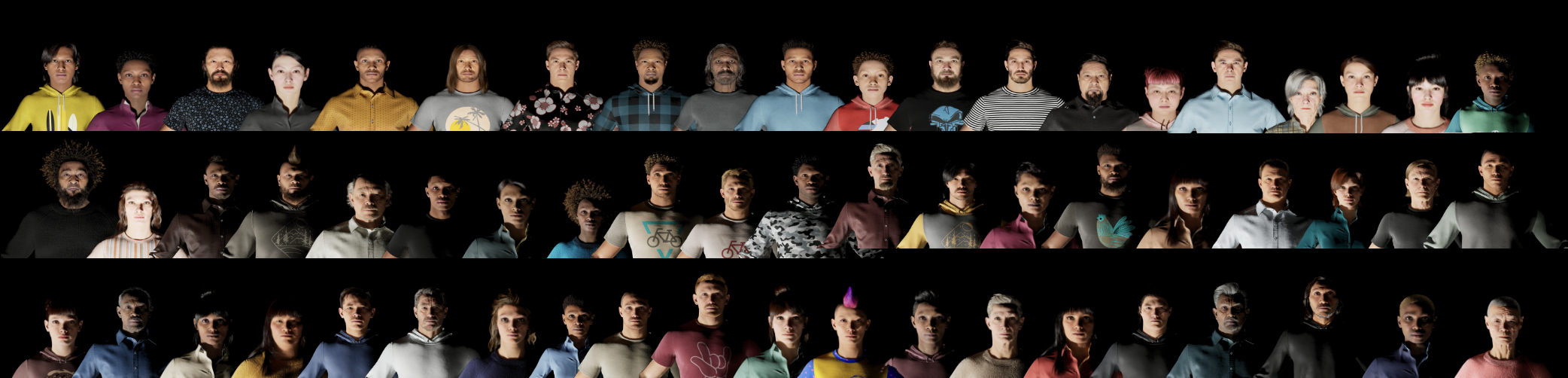}
\end{center}
\vspace{-3mm}
\caption{\rebuttal{Diverse indivisuals in \dataname.
}}
\vspace{-3mm}
\label{fig:individuals}
\end{figure*}

%% file: Tables/tab_lumihuman.tex
\begin{table*}[t]
\caption{Comparison of Openillumination~\cite{liu2024openillumination},DPR~\cite{DPR} and \dataname}
\label{tab:lumihuman}
\vspace{-2mm}
\centering
\resizebox{0.9\textwidth}{!}{
\begin{tabular}{c|c|c|c|c|c|c}
\hline
\textbf{Dataset} & \textbf{Synthesis} & \textbf{Light Positions} & \textbf{Light Movement} & \textbf{Number of Images} & \textbf{Subject} & \textbf{Resolutions} \\ \hline
DPR & 2D & 7 & None & 138K & - & 1024×1024 \\ \hline
Openillumination & Light Stage & 142 & None & 108K & 64 objects & 3000×4096 \\ \hline
\dataname & 3D & \textbf{35,937} & \textbf{>3K}& \textbf{2.3M} & \textbf{65 individuals} & 1024×1024 \\ 
\bottomrule
\end{tabular}
}
\vspace{-2mm}
\end{table*}

%% file: Sections/3-Method.tex
\section{LumiHuman}

Our goal is to achieve unified control over video lighting—a challenging task with significant implications. The primary difficulties can be categorized into three areas:
\textit{(1) Dataset Scarcity}: There is a notable lack of lighting-specific datasets, particularly for videos. Few annotated examples explicitly capture lighting variations, and even fewer provide well-defined lighting information.
\textit{(2) Complexity of Lighting Attributes}: Lighting involves multiple factors, including light source type, direction of illumination, and the material properties of objects. 
Accurately representing lighting effects within the camera's field of view becomes especially important.
\textit{(3) Attribute Decoupling}: Similar to other control tasks, lighting control requires effective decoupling of specific attributes. A major technical challenge lies in isolating lighting information from object appearance in training data, preventing the model from overfitting.

We introduce a portrait lighting dataset, referred to as \dataname.  
\dataname is a continuous lighting video dataset comprising over \textit{220K} different videos (\textit{i.e.,} \textit{2.3 million} images). The resolution of each video is $1024\times1024$.  
\dataname is created using Unreal Engine~\cite{ue} for lighting simulation, allowing for the production of data with known lighting information.
\rebuttal{
As shown in Figure~\ref{fig:individuals}, \dataname includes 65 diverse human subjects, 30K lighting positions, and over 3K lighting trajectories for each people. 
As shown in Table.~\ref{tab:lumihuman}, compared to other lighting datasets Openillumination~\cite{liu2024openillumination} and Deep Portrait Relighting (DPR) dataset~\cite{DPR} (generated from face image dataset Celeb-A~\cite{celeba}), LumiHuman outperforms in light positions, light movements and number of images.}

\input{Figs/fig_lumihuman}

\paragraph{Details}

As shown in Figure~\ref{fig:lumihuman}(a), \dataname can be combined to generate various types of character lighting.
Its 30K lighting positions enable the creation of light and shadow effects across \textit{all areas} of the human face. In Figure~\ref{fig:lumihuman}(b), we present the brightness distribution map for different facial regions. Each ridge in the ridge plot represents a specific facial area, where the horizontal axis indicates brightness and the vertical axis denotes the number of samples corresponding to each brightness level. \dataname comprehensively covers all facial areas and distributes samples across a wide range of brightness levels.
As illustrated in Figure~\ref{fig:lumihuman}(c), we present continuous video frames composed of samples from \dataname. These samples can be flexibly arranged to form diverse lighting trajectories based on user specifications—such as horizontal, vertical, diagonal, arc-shaped, or multi-light-source superpositions.

\paragraph{Lighting Representation}

A straightforward approach to representing lighting effects is to embed lighting parameters as additional input to the model. However, this method requires a large amount of annotated data to establish a reliable mapping between lighting vectors and the two-dimensional image plane.
To better align with the model’s latent space, we propose projecting lighting information into a blank canvas, as illustrated in Figure~\ref{fig:pipeline}(b). For each lighting position, this is visualized as an image in which brighter regions indicate stronger illumination, and darker regions indicate weaker lighting. This representation enables more effective integration of lighting information into the video generation model.

\input{Figs/fig_dataset}

\paragraph{Data Collection}

The \dataname\ collection consists of five key stages:  
\textit{(1) Lighting Design}: As illustrated in Figure~\ref{fig:dataset}(a), we constructed a lighting position matrix—a three-dimensional grid measuring $160\,\text{cm} \times 160\,\text{cm} \times 160\,\text{cm}$. Points within the grid are uniformly spaced at $5\,\text{cm}$ intervals and serve as lighting positions. A point light source moves across these grid points to illuminate the subject from diverse angles.  
\textit{(2) Lighting Trajectory Design}: Within this 3D grid, we defined horizontal, vertical, and diagonal trajectories composed of grid points to simulate a variety of lighting change effects. 
\textit{(3) Character Construction}: To generate high-quality portrait lighting data, we employed the MetaHuman dataset \cite{metahuman}, which provides 3D models of 65 diverse individuals. This variation enhances the richness of visual lighting effects across different characters.  
\textit{(4) Flexibility and Storage Optimization}: To support a wide range of lighting path variations while managing storage demands—particularly due to duplicate frames from identical lighting positions—we provide a flexible, lightweight image-video dataset. The dataset includes images rendered from different lighting positions within the 3D grid, with each character associated with $33 \times 33 \times 33$ unique samples. Videos can be generated using predefined lighting trajectories, or users can define new paths to simulate additional lighting effects, as demonstrated in Figure~\ref{fig:dataset}(b).  
\textit{(5) Text Annotation and Augmentation}: For automated video captioning, we utilized BLIP \cite{li2022blip}. To enrich the contextual diversity of the dataset, we further applied GPT-4 \cite{achiam2023gpt} for caption augmentation, generating a wide range of descriptive narratives, as illustrated in Figures~\ref{fig:dataset}(c) and~(d).

\input{Figs/fig_pipeline}

\section{LumiSculpt}

\subsection{Integrating Lighting into Video Generators}

\dataname enables lighting to be represented in pixel space and parameterized as an input to standard visual models. We extract lighting features using a Variational Autoencoder (VAE), shared with the generative model, and feed them into a dedicated lighting encoder. As illustrated in Figure~\ref{fig:pipeline}(b), the lighting encoder adopts a transformer architecture composed of self-attention layers, enabling it to compute global attention scores across video frames. This design effectively captures the spatial and temporal dynamics of lighting throughout a video clip.

The lighting encoder takes the lighting features as input and passes them through transformer blocks that match the number of layers in the backbone model. These blocks output a sequence of latent representations with dimensions aligned to those of the backbone model, enabling seamless feature fusion. Our objective is to integrate these latents into the DiT architecture of the T2V model. Specifically, the latent video features, denoted \(z_t\), are combined with the lighting features \(c_t\) via element-wise addition. The resulting features are then passed through a linear layer to produce the output for the next layer, with a guidance scale hyperparameter set to 0.5.

\rebuttal{As shown in Figure~\ref{fig:pipeline}(d), \sysname employs a 3D self-attention mechanism in the lighting encoder and utilizes multi-stage weighting as the conditional injection mechanism. In comparison, ControlNet uses a U-Net encoder to extract features and injects conditions via additive latent fusion.}

\subsection{Lighting Learning}

We adopt a data-driven approach to learn complex lighting effects, specifically capturing the impact of light projection from various positions on the human face. This approach is implemented using \dataname. However, a key challenge arises when learning lighting directly from data rendered in Unreal Engine: the leakage of appearance information. This issue stems from the dataset’s consistent backgrounds and layouts, which, while beneficial for stable training, also increase the risk of the model overfitting to specific appearances.

This highlights a central challenge in lighting control: how can lighting be disentangled from other visual attributes?  
We address this by proposing a lighting-appearance disentanglement method, which incorporates a dual-branch architecture alongside a novel disentanglement loss function. This framework enables the model to isolate lighting information while mitigating the influence of appearance-related features.

\paragraph{Dual-Branch Framework}

A straightforward approach to obtaining diverse appearance data is to use additional video datasets as regular samples, thereby exposing the model to a wider range of appearances. However, acquiring large-scale, diverse data with known lighting conditions remains challenging. To address this, we generate regular samples using generative models.

As illustrated in Figure~\ref{fig:pipeline}(c), we propose a dual-branch framework consisting of a training branch and a frozen reference branch. The frozen branch utilizes a pre-trained foundational denoising model to serve as an appearance reference. During training, both branches receive the same textual condition and noisy latent input, producing predicted noise values \(\epsilon_t\) and \(\epsilon_t^{reg}\), respectively. The diverse appearance priors from the pre-trained model are captured in \(\epsilon_t^{reg}\), enabling the generation of regular samples in a cost-efficient manner.

\input{Tables/tab_comparison}

\paragraph{Loss Functions}
We train the model to capture the overall distribution of the dataset using a denoising loss \(\mathcal{L}_{\text{denoise}}\), which captures both lighting and appearance information. However, since appearance variation is not required, we propose a disentanglement loss, \(\mathcal{L}_{\text{dis}}\), to evaluate and suppress appearance information.
\(\mathcal{L}_{\text{dis}}\) satisfy two criteria: (1) it should effectively quantify the consistency of appearance between the latent representations of two videos, and (2) it should be invariant to planar spatial information, as dependence on pixel location could hinder the learning of lighting distributions.
Inspired by neural style transfer~\cite{Jing:2020:NST}, which similarly aims to preserve style while discarding structural information, 
we adopt Adaptive Instance Normalization (AdaIN)~\cite{huang2017arbitrary} to compare the feature statistics—mean and variance—of the latents from two branches. This effectively captures the distribution of appearance features. By aligning the appearance distribution of the training branch with that of a frozen reference branch, we retain the diverse generative capacity of the original model while filtering out redundant appearance signals.
Under this framework, the generative model is optimized to match the training video via \(\mathcal{L}_{\text{denoise}}\), while \(\mathcal{L}_{\text{dis}}\) enforces appearance disentanglement. Together, they enable precise lighting control while preserving generalization.
The overall training objective is defined as:\
\begin{equation}
\begin{aligned}
\mathcal{L}_{\text{dis}} &= \left\| \sigma(z^{\text{pred}}_0) - \sigma(z^{\text{reg}}_0) \right\|_2 + \left\| \mu(z^{\text{pred}}_0) - \mu(z^{\text{reg}}_0) \right\|_2, \\
\mathcal{L}_{\text{denoise}} &= \mathbb{E}_{z_{1:N}, \epsilon, c_t, t} \left[ \left\| \hat{\epsilon}(z^{\text{pred}}_{1:N}, c_t, t) - \epsilon \right\|^2 \right], \\
\mathcal{L}_{\text{total}} &= \mathcal{L}_{\text{denoise}} + \beta \mathcal{L}_{\text{dis}},
\end{aligned}
\end{equation}
where \(\sigma(\cdot)\) and \(\mu(\cdot)\) denote the standard deviation and mean, respectively. \(z_0^{\text{reg}} = z_t^{\text{reg}} - \epsilon_{\text{reg}}\) and \(z_0^{\text{pred}} = z_t^{\text{pred}} - \epsilon_{\text{pred}}\) represent the predicted denoised latents of the frozen and training branches at timestep \(t\). \(N\) is the total number of denoising steps, \(c_t\) is the textual condition, and \(\beta\) is a balancing coefficient set to 3.0.

\input{Figs/fig_comparison}

%% file: Figs/fig_lumihuman.tex
\begin{figure*}[t]
\begin{center}
\includegraphics[width=0.99\linewidth]{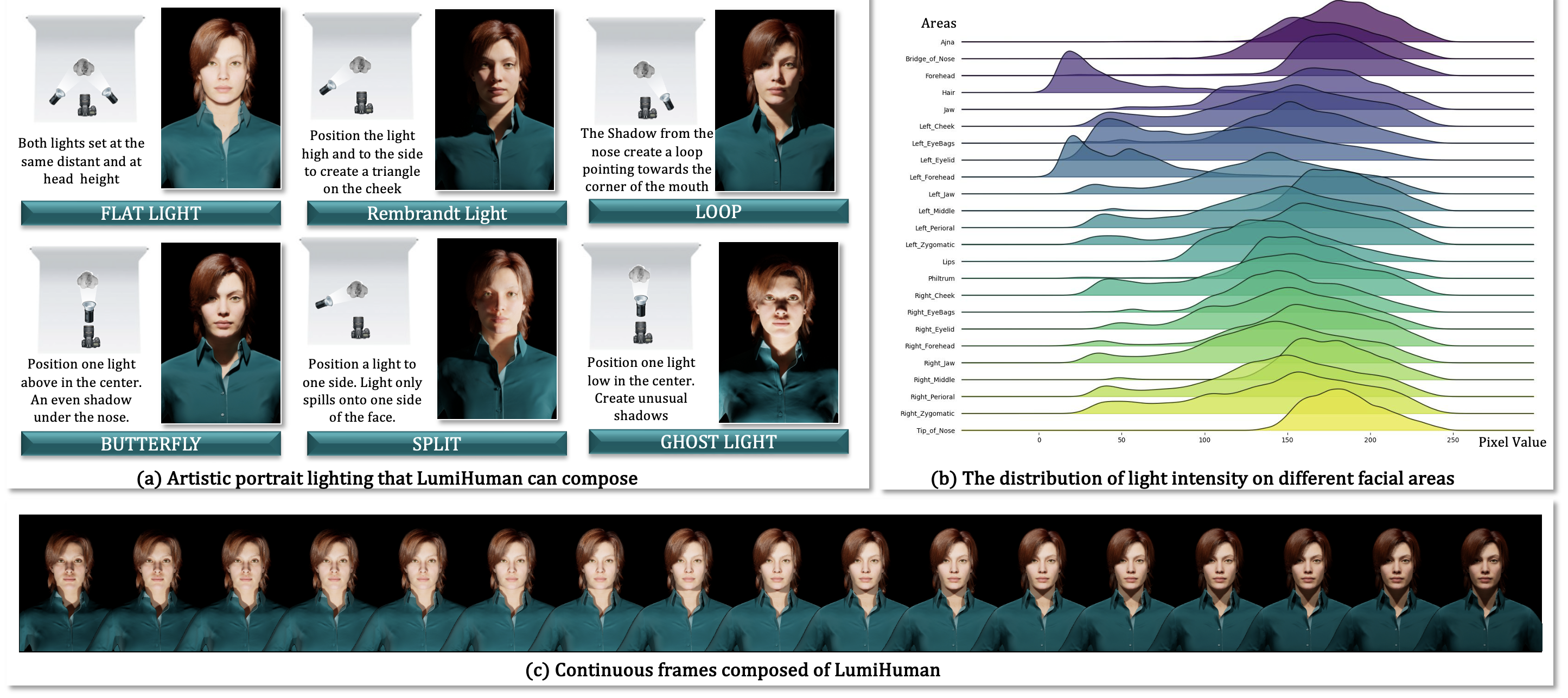}
\end{center}
\vspace{-3mm}
\caption{\rebuttal{\textbf{(a)} \dataname offers a variety of basic elements that can be combined to form various types of portrait lighting, widely applicable to a range of tasks related to character lighting. 
\textbf{(b)} shows the distribution of light intensity on different facial areas of the characters; \dataname's lighting matrix can cover all areas of the face and produce a significant range of light and shadow variations. \textbf{(c)} shows an example of creating a continuous lighting video using \dataname.}
}
\label{fig:lumihuman}
\end{figure*}

%% file: Figs/fig_dataset.tex
\begin{figure*}[t]
\begin{center}
\vspace{-2mm}
\includegraphics[width=0.99\linewidth]{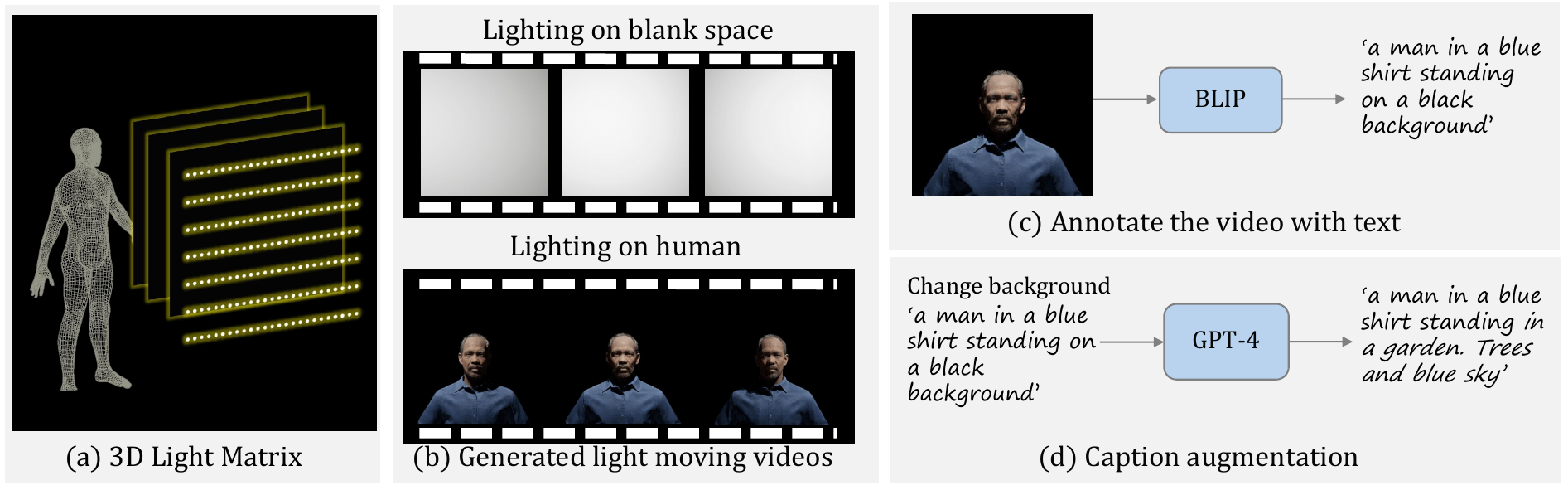}
\end{center}
\vspace{-3mm}
\caption{The collection process of \dataname includes: (a) designing a 3D point light source matrix of $33\times33\times33$ lighting points, (b) rendering single-frame images and generating portrait lighting videos with various path lighting and lighting reference videos, (c) annotating with a BLIP model, and (d) producing enhanced background captions using a large language model.
}
\vspace{-3mm}
\label{fig:dataset}
\end{figure*}

%% file: Figs/fig_pipeline.tex
\begin{figure*}[t]
\begin{center}
\includegraphics[width=0.99\linewidth]{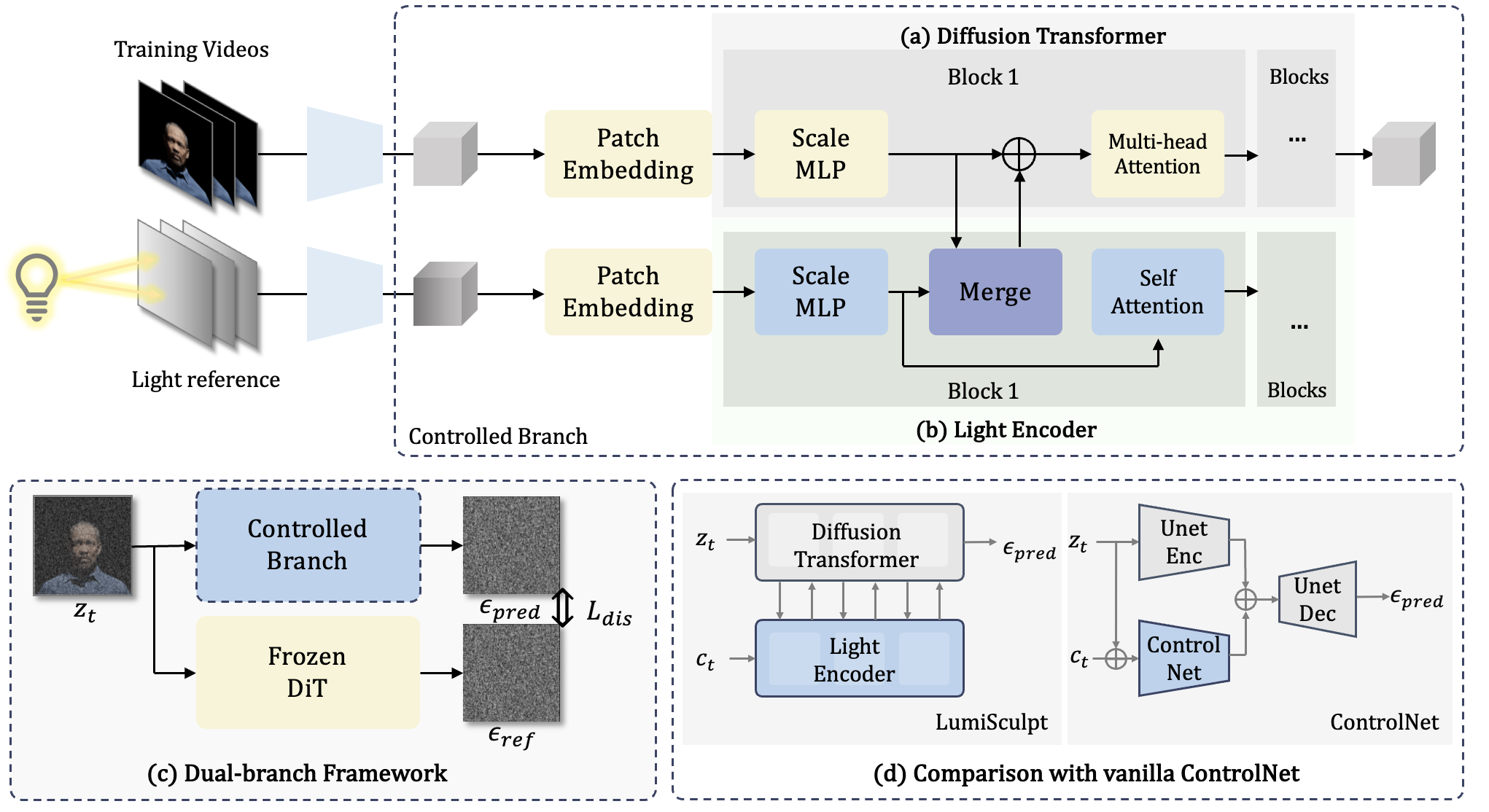}
\end{center}
\caption{ 
The pipeline of \sysname consists of the generation backbone, i.e. the controlled branch, which includes (a) a diffusion transformer (DiT), a pre-trained video denoising network, and (b) a light encoder, a trainable external transformer network. The light encoder takes light reference latents as input and processes them through various blocks to produce a light condition sequence. 
This sequence is integrated into the generation backbone using several merge modules within each block. During training, we propose a \textbf{(c) dual-branch framework} including a controlled branch and a frozen branch, which provide regularization for diverse appearances. The frozen branch is a DiT with frozen parameters, sharing weights with (a). Both branches predict noise, resulting in $\epsilon_{pred}$ and $\epsilon_{reg}$, which are used to compute the disentanglement loss $\loss_{dis}$.
\rebuttal{(c) and (d) show that \sysname differs form ControlNet~\citep{controlnet} in terms of model structure, condition injection, training manners and objectives.}}
\label{fig:pipeline}
\end{figure*}

%% file: Tables/tab_comparison.tex
\begin{table}[th]
\caption{Quantitative experimental results and ablation study results.The best results are marked as \textbf{bold} and the seconds one are marked by \underline{underline}.}
\label{tab:comparison}
\begin{center}
\vspace{-3mm}
\resizebox{1\linewidth}{!}{
\begin{tabular}{c|cc|cc|c}
\toprule
\multirow{2}{*}{Method}  &\multicolumn{2}{c|}{Consistency}&\multicolumn{2}{c|}{Lighting Accuracy}&\multicolumn{1}{c}{Quality}
\\ \cline{2-6}
& CLIP$\uparrow$ & LPIPS$\downarrow$ & Direction$\downarrow$ & Brightness$\uparrow$ & CLIP$\uparrow$ \\
\hline
Open-Sora & 0.9845 & 1.3503  & 0.4542 & 0.8229 & 0.3182  \\ 
IC-Light & 0.9703 & 2.5329 & 0.5264 & 0.8632 & 0.3145 \\ 
\rebuttal{ControlNet} & 0.8081 & 5.9324 & 0.5500 & 0.8032 & \underline{0.3440} \\ 
\hline
Ours(full model) & \underline{0.9951} & 1.1312 & 0.3500 & \underline{0.8779} & \textbf{0.3597} \\ 
Ours(w/o caption aug) & 0.9948 & \underline{1.1211}  & \underline{0.2992} & \textbf{0.9269} & 0.3416 \\
Ours(w/o $\loss_{dis}$) & \textbf{0.9957} & \textbf{1.1033} & \textbf{0.1945} & 0.8363 & 0.2909 \\
\bottomrule
\end{tabular}
}
\vspace{-2mm}
\end{center}
\end{table}

%% file: Figs/fig_comparison.tex
\begin{figure*}[t]
\begin{center}
\includegraphics[width=0.9\linewidth]{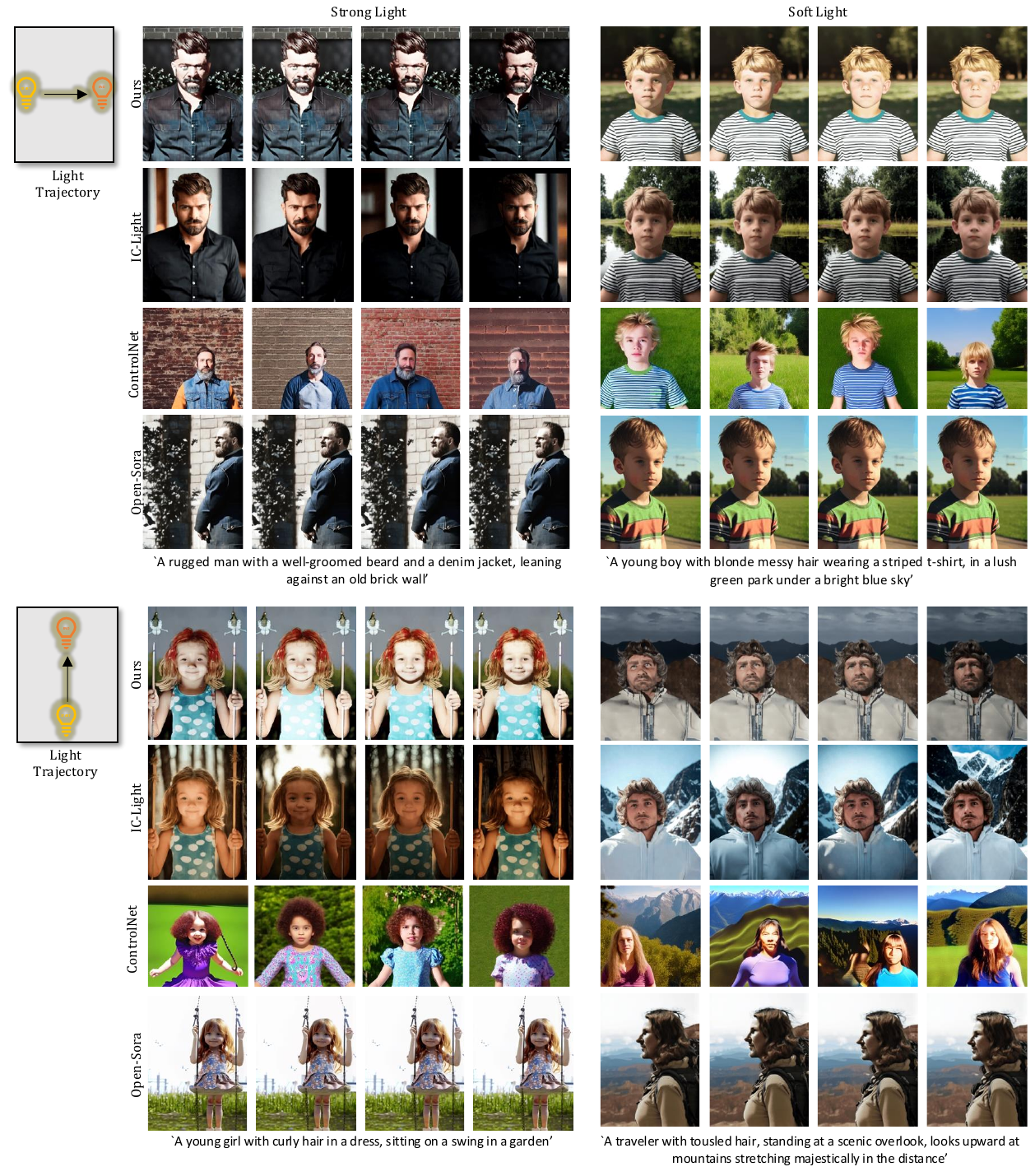}
\end{center}
\vspace{-5mm}
\caption{Comparison results with state-of-the-art methods IC-Light~\citep{iclight}, \rebuttal{ControlNet}~\citep{controlnet} and Open-Sora~\citep{opensora}. The classic horizontal and vertical directions for light movement and two brightness levels are tested to achieve a comprehensive qualitative evaluation.
}
\vspace{-5mm}
\label{fig:comparison}
\end{figure*}

%% file: Sections/4-Exp.tex
\section{Experiments}
\subsection{Experimental Setup}

\paragraph*{Methods for comparison~}
We compare our approach with state-of-the-art text-to-video generation methods Open-Sora~\cite{opensora}, image relighting method IC-light~\cite{iclight}, \rebuttal{and image control method ControlNet~\cite{controlnet}}.
\paragraph*{Metrics~}
We employ a variety of quantitative and qualitative metrics to assess the lighting accuracy, inter-frame coherence, and visual-text similarity of generated videos.
\paragraph*{Evaluation dataset~}
We use 500 different light paths and captions not present in the training dataset as conditions to guide the comparative methods in generating evaluation videos.
\paragraph*{Implementation details.}
In all video generation experiments, we use Open-Sora v1.2.0~\cite{opensora} with the default network architecture.
We set a learning rate of $1\times 10^{-4}$.
The input video resolution is $640 \times 480 \times 29$.
The training process for each motion requires approximately $800 \sim 1500$ iterations using eight NVIDIA A100.
The number of inference steps is set to $T = 50$ and the guidance scale is set to $w = 7.5$.

\subsection{Quantitative Evaluations}

Table~\ref{tab:comparison} presents five quantitative metrics used for evaluation:
(1) \textit{Frame-wise CLIP image similarity} assesses semantic-level video coherence by measuring the similarity of frame-wise CLIP~\cite{clip} image embeddings. A higher value indicates stronger inter-frame similarity and better semantic stability in the generated video.
(2) \textit{Frame-wise LPIPS consistency} evaluates feature-level coherence by calculating frame-wise LPIPS. A lower value reflects smaller feature discrepancies, indicating higher inter-frame consistency.
(3) \textit{Lighting direction RMSE} is computed for each frame and assess the consistency of the generated video’s lighting direction with the reference. A smaller RMSE indicates better alignment with the target lighting direction.
(4) \textit{Brightness consistency} involves segmenting each video frame into patches and computing the average brightness for each patch to construct a brightness distribution. This distribution is used to measure the consistency between the generated video and the reference, independent of absolute brightness values.
(5) \textit{CLIP text-image similarity} measures the alignment between the generated video frames’ CLIP image embeddings and the text embedding of the caption. A higher similarity indicates better generation quality.
As shown in the results, \sysname outperforms Open-Sora-Plan~\cite{opensora} and IC-Light~\cite{iclight} by maintaining strong inter-frame consistency and text-image alignment, while also achieving precise lighting control.

\subsection{Qualitative Evaluations}

As shown in Figure~\ref{fig:comparison}, due to the absence of video illumination control methods, we compare our approach with the image illumination control method IC-Light based on diffusion models, and the video generation method Open-Sora. We consider two light intensity levels, strong and soft, as well as horizontal and vertical lighting movement directions. Since IC-Light is designed for relighting existing images, we use portraits generated by our method as foreground guidance.
IC-Light is capable of producing single-frame images with accurate lighting directions, but due to a lack of inter-frame awareness, the coherence of the output video is poor, with noticeable flickering in the background. 
Open-Sora can generate coherent and aesthetically pleasing videos, but struggles to control lighting direction via textual conditions, resulting in relatively unchanged lighting throughout the video. 
Our method not only ensures video coherence and visual quality but also achieves precise control over lighting trajectory and intensity.
Video results are provided in the supplementary material.

\input{Figs/fig_ablation}
\input{Tables/tab_scale}

\subsection{Ablation Study}

As shown in the $1^{st} \sim 3^{rd}$ and last rows of Figure~\ref{fig:ablation}, we present the results of ablating different modules of \sysname. 
Removing caption augmentation from \dataname leads to a lack of diverse textual guidance, causing the model during training to rely solely on text conditions that exactly match the dataset, thus improving appearance fitting. As shown in the second row, the generated results exhibit consistent pose and layout. Without the dual-branch structure and decoupling loss, as shown in the third row, the generated appearances tend to overfit the training data, making it challenging to produce diverse backgrounds. 
As illustrated in the first row, the complete \sysname successfully balances diverse appearances with accurate lighting.  
As shown in the $4^{th} \sim 7^{th}$ rows of Figure~\ref{fig:ablation} and Table~\ref{tab:scale}, we present both qualitative and quantitative analysis results of varying the hyper-parameter \textit{guidance scale}. During inference, the standard \sysname sets the guidance scale to 0.5. Increasing the guidance scale intensifies the strength of lighting guidance, enhancing the accuracy of lighting direction and brightness, but an excessively large guidance scale can undermine the model's ability to generate diverse outputs. In contrast, decreasing the guidance scale reduces the strength of lighting guidance, leading to a decrease in lighting accuracy.

\paragraph*{Generalization}

\rebuttal{As shown in the Figure~\ref{fig:animals}, \sysname provides lighting priors on animals, which demonstrates the generalization ability to real-world cases.
}

\input{Figs/fig_animals}

%% file: Figs/fig_ablation.tex
\begin{figure*}[t]
\begin{center}
\vspace{-5mm}
\includegraphics[width=0.9\linewidth]{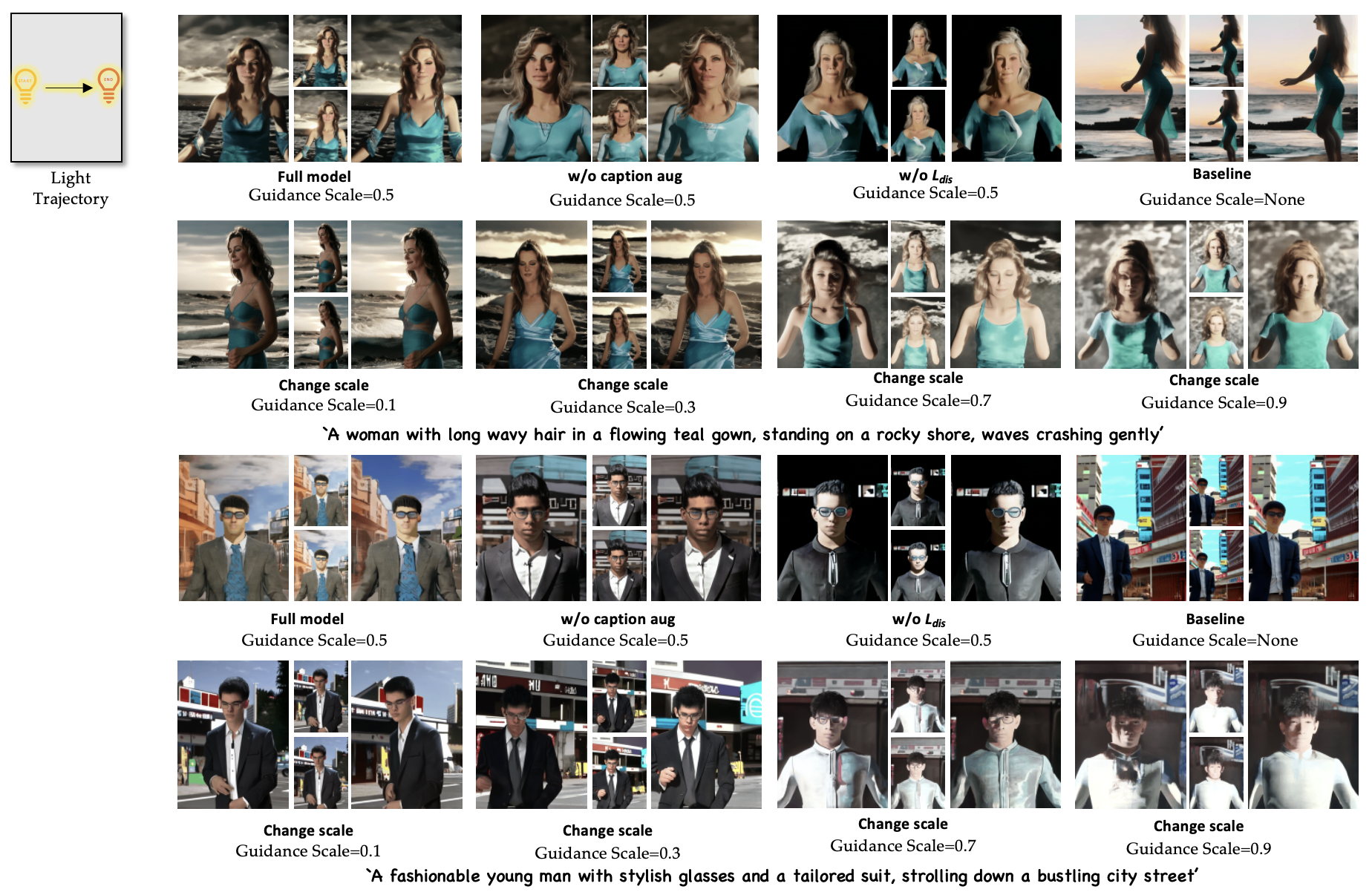}
\end{center}
\vspace{-5mm}
\caption{Ablation study results. We conducted ablations on key components related to model training, including caption augmentation and $\loss_{dis}$, both of which contribute to the diversity of generated content. We presented results with varying values of the important parameter \textit{guidance scale}, which affects the accuracy of lighting and the diversity of appearance, related to model inference.
}
\vspace{-3mm}
\label{fig:ablation}
\end{figure*}

%% file: Tables/tab_scale.tex
\begin{table}[t]
\caption{Experimental results of hyper-parameter \textit{guidance scale}. The best results are marked as \textbf{bold} and the seconds one are marked by \underline{underline}}
\vspace{-3mm}
\label{tab:scale}
\begin{center}
\begin{tabular}{c|c|c|c}
\toprule
\multicolumn{1}{c|}{\bf Scale}  &\multicolumn{1}{|c|}{\bf Consistency$\uparrow$}&\multicolumn{1}{|c|}{\bf Accuracy$\downarrow$}&\multicolumn{1}{|c}{\bf Quality$\uparrow$} \\
\hline
\textit{scale}=0.1 & 0.9943 & 0.4239 & 0.3163 \\ 
\textit{scale}=0.3 & \textbf{0.9964} & 0.3825 & \textbf{0.3814} \\ 
\textit{scale}=0.5 & \underline{0.9951} & 0.3500 & \underline{0.3597} \\ 
\textit{scale}=0.7 & 0.9939 & \textbf{0.2484} & 0.3499 \\ 
\textit{scale}=0.9 & 0.9902 & \underline{0.2922} & 0.2941 \\
\bottomrule
\end{tabular}
\vspace{-5mm}
\end{center}
\end{table}

%% file: Figs/fig_animals.tex
\begin{figure}[t]
\begin{center}
\includegraphics[width=0.99\linewidth]{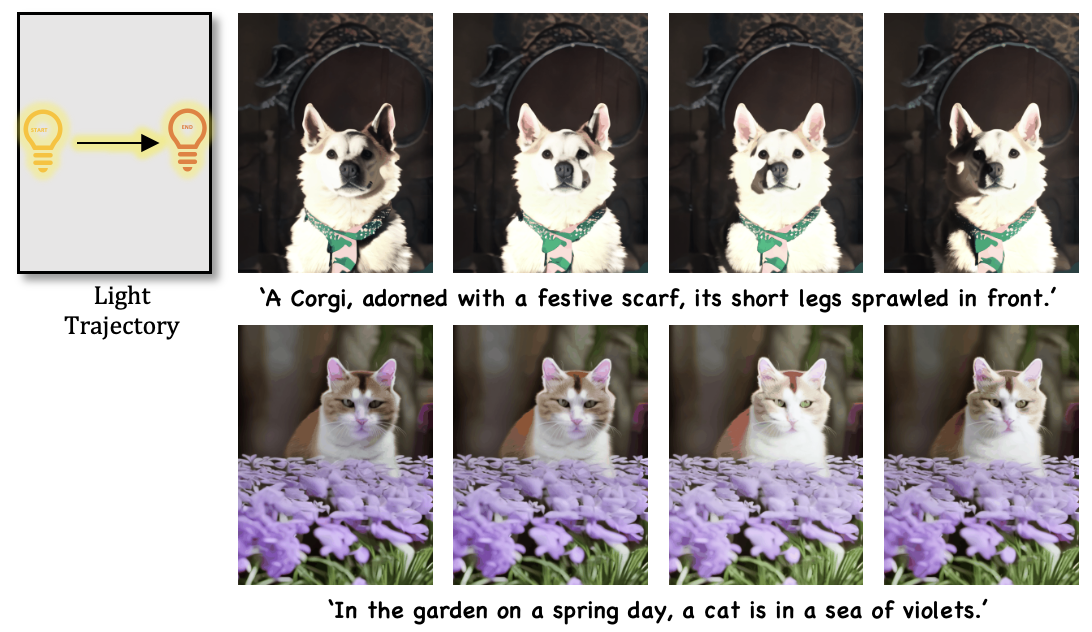}
\end{center}
\vspace{-5mm}
\caption{\rebuttal{The results of \sysname on animals. 
}}
\vspace{-5mm}
\label{fig:animals}
\end{figure}

%% file: Sections/5-Conclusion.tex
\section{Conclusion}

In this paper, we address the challenge of lighting control in text-to-video generation. To address data scarcity, lighting representation difficulties, and lighting injection complexities, we introduce a flexible portrait lighting dataset, \dataname, along with a plug-and-play lighting guidance method, \sysname.
Our proposed dual-branch structure and associated loss function for decoupling are not only effective for lighting control but also have the potential to be generalized across a variety of generative tasks. As video generation techniques continue to evolve, enabling precise lighting control becomes an increasingly important research direction, driven by significant aesthetic demands from both professionals and the general public. We believe that \dataname and \sysname will serve as valuable resources and methodologies for future explorations in this domain.